\RequirePackage{amsmath}
\documentclass[runningheads]{llncs}
\usepackage{graphicx}
\usepackage{amsmath,amssymb} 
\usepackage{color}
\usepackage[caption=false]{subfig}

\begin{document}
\title{Inner Space Preserving Generative Pose Machine} 

\titlerunning{Inner Space Preserving Generative Pose Machine}
%
\author{Shuangjun Liu \and Sarah Ostadabbas}
%
\authorrunning{S. Liu and S. Ostadabbas}
%

\institute{Augmented Cognition Lab, Electrical and Computer Engineering Department,\\
	Northeastern University, Boston, USA\\
\email{\{shuliu,ostadabbas\}@ece.neu.edu}\\
\url{http://www.northeastern.edu/ostadabbas/}}

\newcommand{\figWillamsRushB}{
\begin{figure}[t]
 \centering
 \includegraphics[width=.85\textwidth]{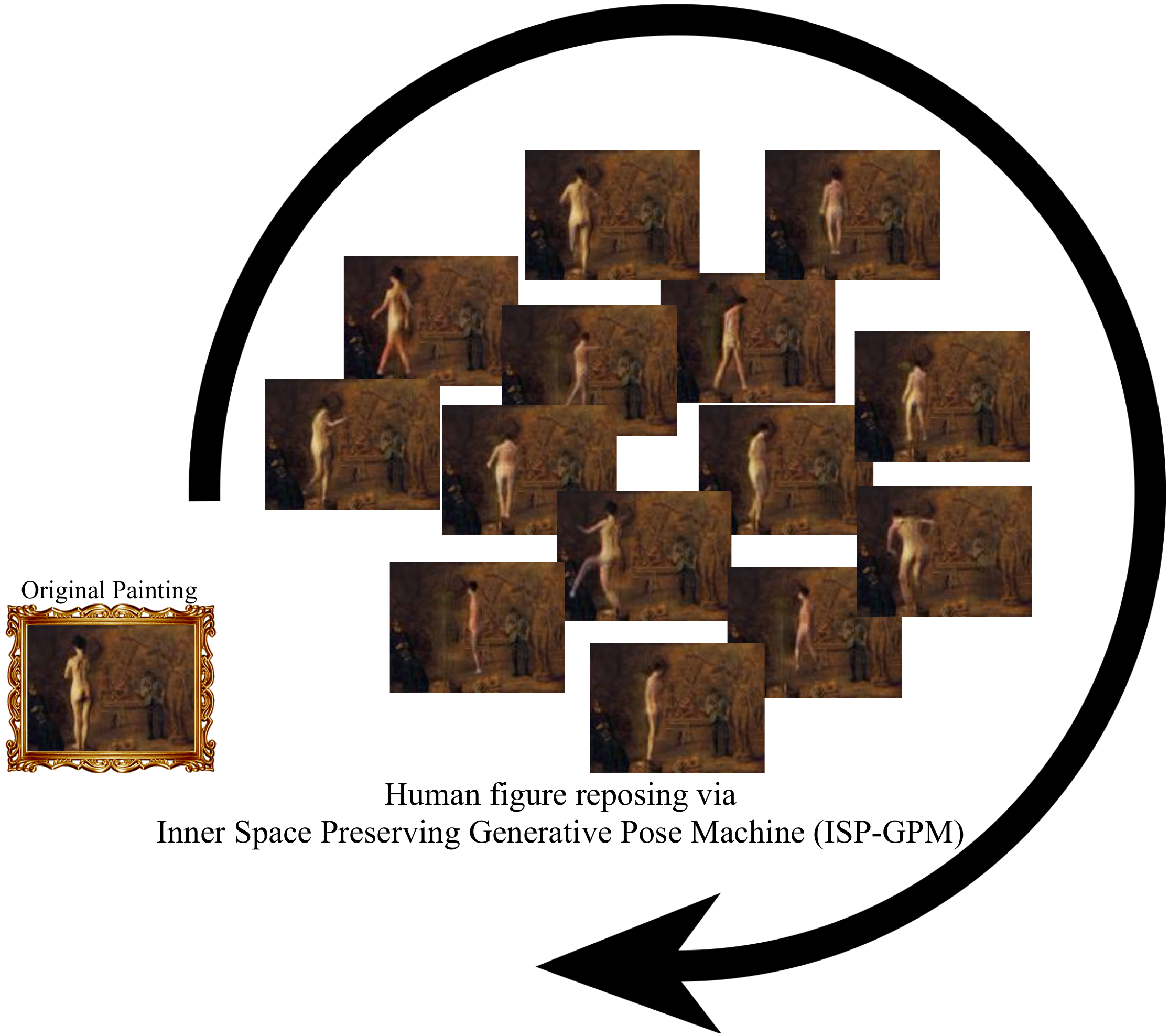}
 \caption{Inner space preserving reposing of one of Thomas Eakins' paintings: William Rush Carving His Allegorical Figure of the Schuylkill River, 1908.}
 \vspace{-.3in}
\label{fig:WillamsRushB}
\end{figure}
}

\newcommand{\figNipsLWW}
{
\begin{figure}[t]
 \centering
 \subfloat[]{
 \label{fig:cub_shrink_1_black}
 \includegraphics[width=0.48\linewidth,{trim=0in 0in 0in 0in,clip=true}]{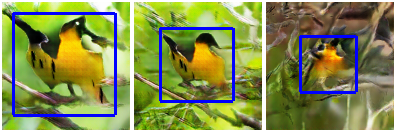}}
 \subfloat[]{\label{fig:cub_shrink_1_red}\includegraphics[width=0.48\linewidth,{trim=0in 0in 0in 0in,
  clip=true}]{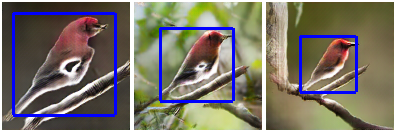}} 
\caption{Generated bird figures from work presented in \cite{reed2016learning} with captions as: (a) this bird has a \emph{black} head, a pointy orange beak, and yellow body, (b) this bird has a \emph{red} head, a pointy orange beak, and yellow body.}
\label{fig:NipsLWW}
\vspace{-.25in}
\end{figure}
}

\newcommand{\figGPMframe}{
\begin{figure}[t]
 \centering
 \includegraphics[width=0.95\linewidth,{trim=0in 0in 0in 0in,
  clip=true}]{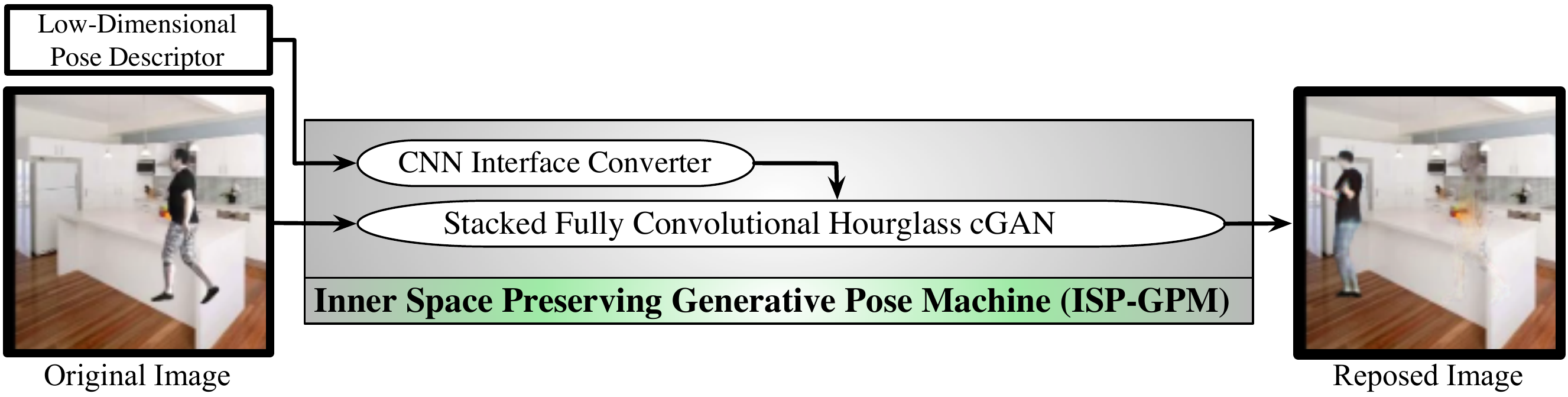}
 \caption{An overview of the Inner Space Preserving Generative Pose Machine (ISP-GPM) framework.}
  \vspace{-.3in}
\label{fig:GPMframe}
\end{figure}
}

\newcommand{\figFChourglassCGAN}{
\begin{figure}[t]
 \centering
 \includegraphics[width=1\linewidth,{trim=0in 0in 0in 0in,
  clip=true}]{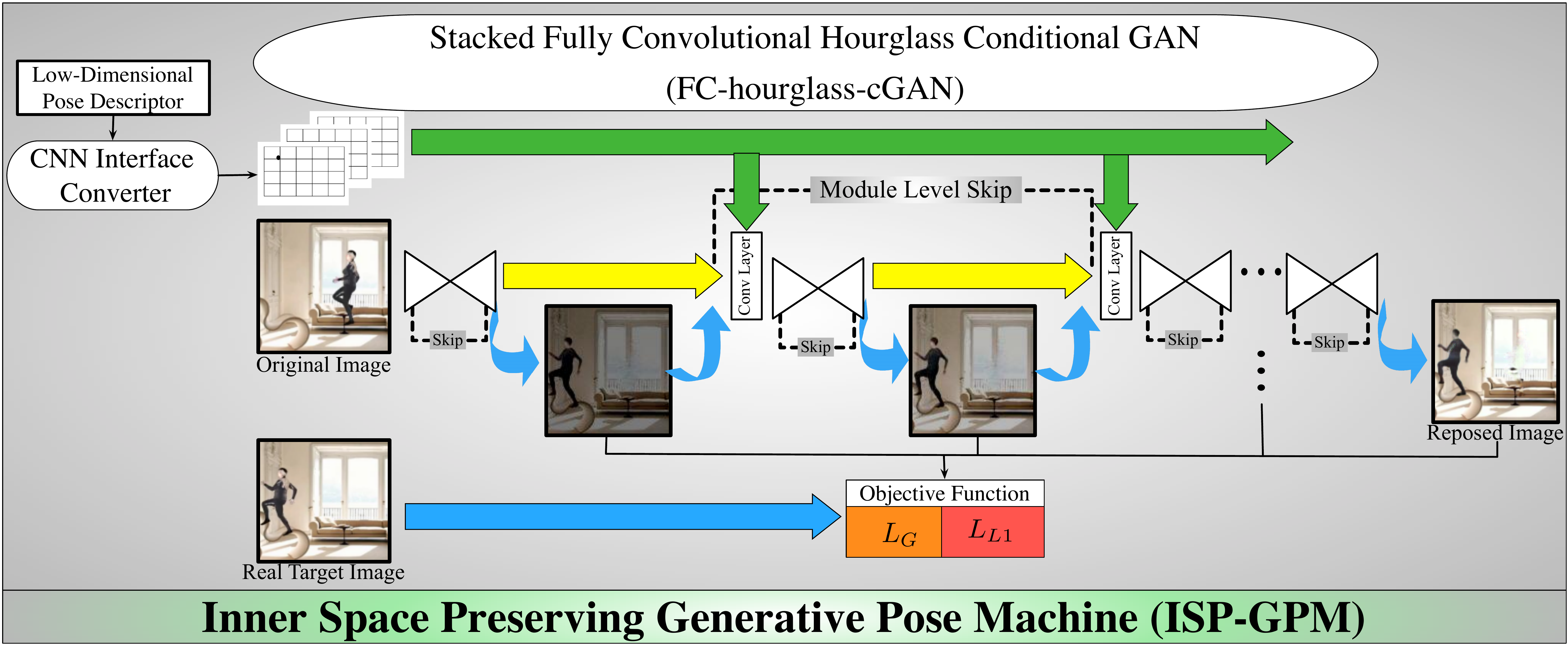}
 \caption{Inside the stacked FC-hourglass-cGAN part of the ISP-GPM. Blue arrows stand for the image flow, yellow arrows for the hourglass feature maps, and green arrows  for $J_{Map}$ flow.}
\label{fig:FChourglassCGAN}
  \vspace{-.3in}
\end{figure}
}
\newcommand{\figFChourglass}{
\begin{figure}[h]
 \centering
 \includegraphics[width=0.84\linewidth,{trim=0in 0in 0in 0in,
  clip=true}]{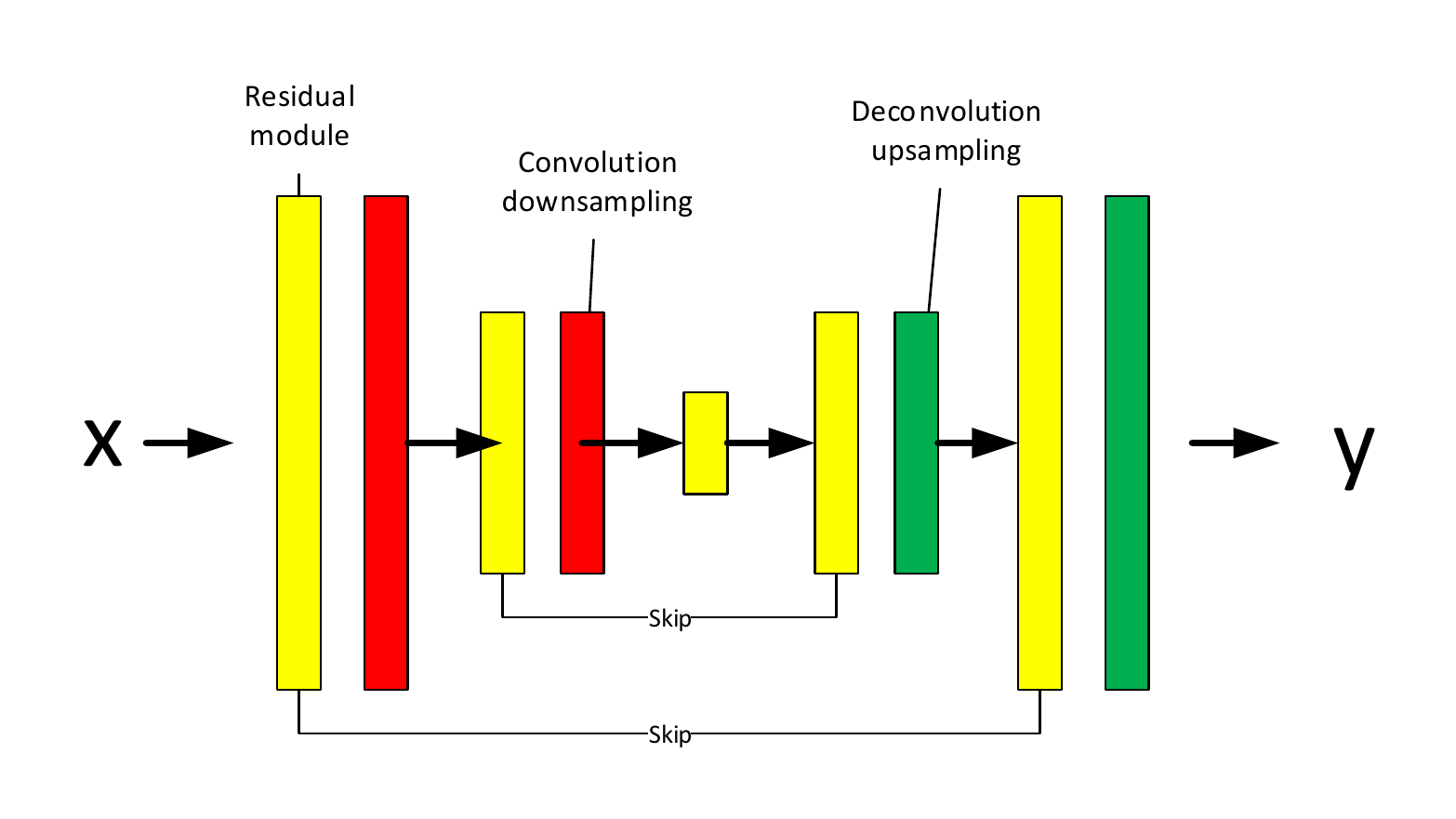}
 \caption{Fully convolutional conditional hourglass. Yellow stands for Residual module \cite{he2016deep}, red stands for convolutional donwsampling process which includes batchNorm, ReLU and convolutional layer, green stands for deconvolutional upsampling process which includes batchNorm, ReLU and deconvolutional layer. The actual downsample/upsample level depends on the parameter which is 3 here for this demo case.}
\label{fig:FChourglass}
\vspace{-.2in}
\end{figure}
}

\newcommand{\figDlayersLabs}{
\begin{figure}[t]
 \centering
 \includegraphics[width=0.84\linewidth,{trim=0in 0in 0in 0in,
  clip=true}]{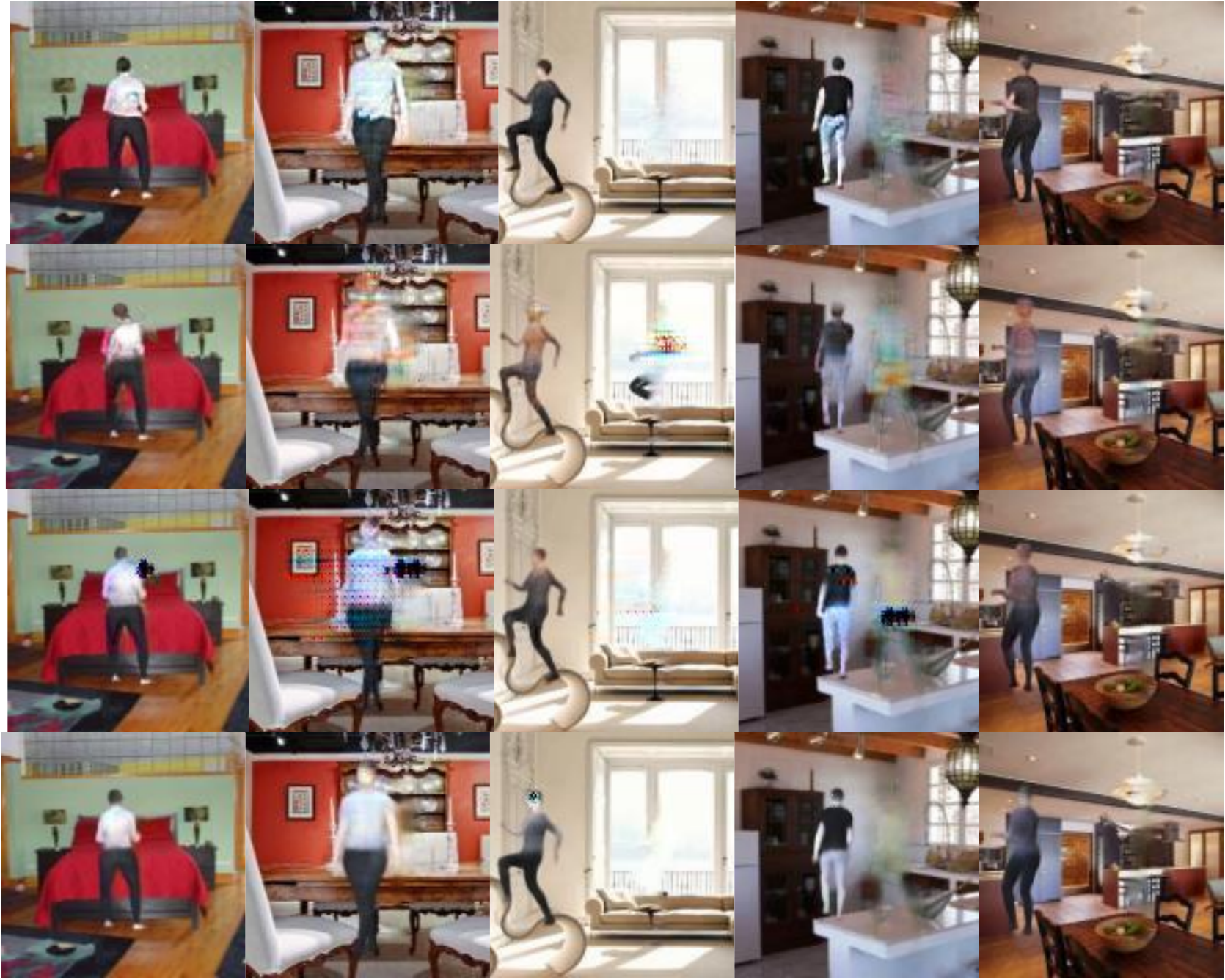}
 \caption{Reposed human figure under different network configurations: 1st to 3rd row with two to four layers discriminator network and 4th row without discriminator but only L1 loss.}
 \vspace{-.2in}
\label{fig:DlayersLabs}
\end{figure}
}

\newcommand{\figArtsRps}{
\begin{figure}[t]
 \centering
 \includegraphics[width=0.84\linewidth,{trim=0in 0in 0in 0in,
  clip=true}]{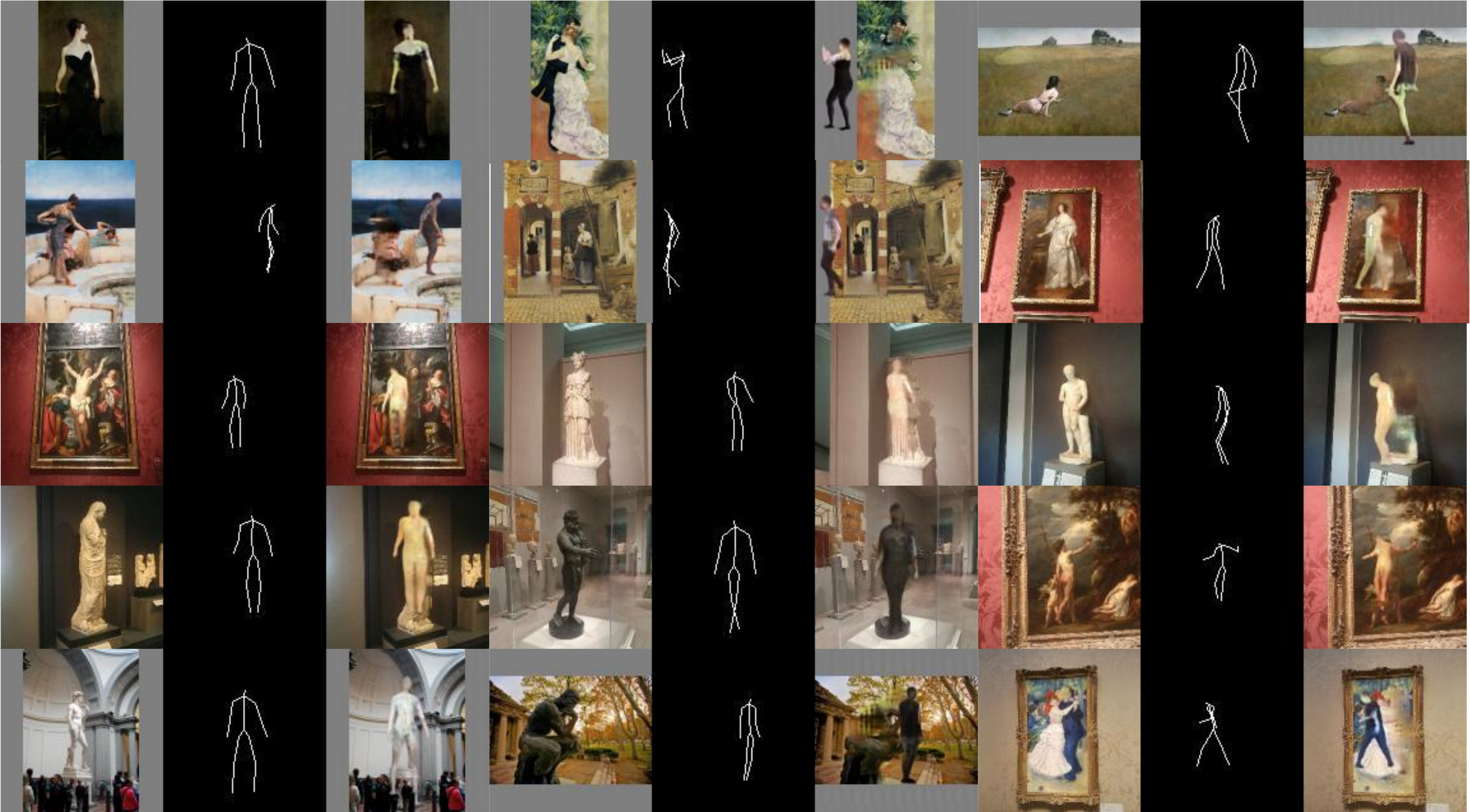}
 \caption{ISP reposing of art works, in left to right, top to bottom order: 
 Madame X (1884)--John Singer Sargent, 
 Dance at Bougival (1882-1883)--Pierre Auguste Renoir, Christinas World--Andrew Wyeth, 
 Silver Favourites (1903)--Lawrence Alma-Tadema, 
 The Courtyard of a House in Delft (1658)--Pieter de Hooch, Isabella Lady de La Warr--Anthony van Dyck, 
 Saint Sebastian Tended--Saint Irene and her Maid-Bernardo Strozzi,
 Athena Parthenos (Roman, 2nd or 3rd century, Marble),
 Athlete with a Scraper (Roman, Imperial period, Marble), Woman from a Funerary Naiskos  (Greek, late classical period, Marble), 
 Hercules (Roman, early Imperial period, Bronze), Bacchus discovering Ariadne--Jacob jordaens, 
 David full body-- Michelangelo, 
 The thinker--Auguste Rodin, 
 Dance at Bougival--Pierre Auguste Renoir.}
  \vspace{-.3in}
\label{fig:artsRps}
\end{figure}
}


\newcommand{\figRealRpsB}
{
\begin{figure}[t]
 \centering
 \subfloat[]{
 \label{fig:mpiiRps_2}
 \includegraphics[width=0.3\linewidth,{trim=0in 0in 0in 0in,clip=true}]{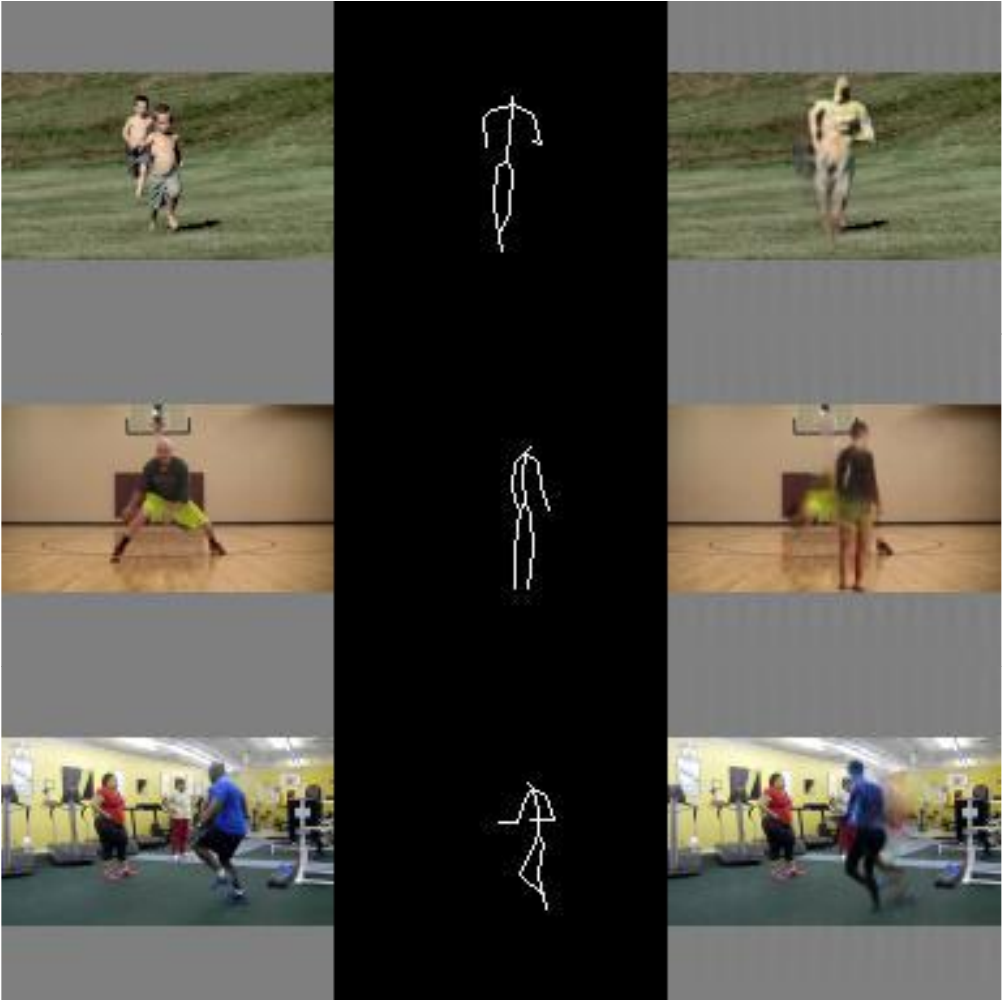}}
  \subfloat[]{\label{fig:lspRps_2}
 \includegraphics[width=0.3\linewidth,{trim=0in 0in 0in 0in,
  clip=true}]{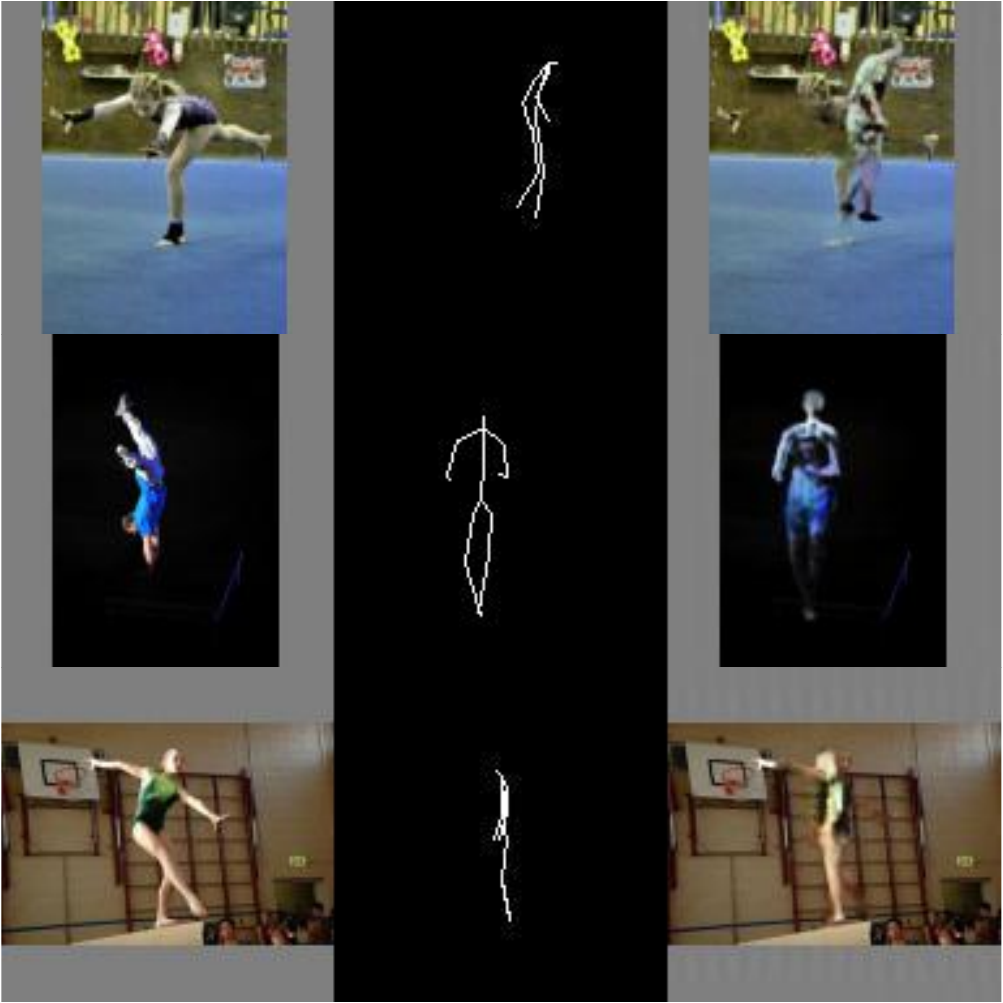}} 
 \subfloat[]{\label{fig:artsRps_2}
 \includegraphics[width=0.3\linewidth,{trim=0in 0in 0in 0in,
  clip=true}]{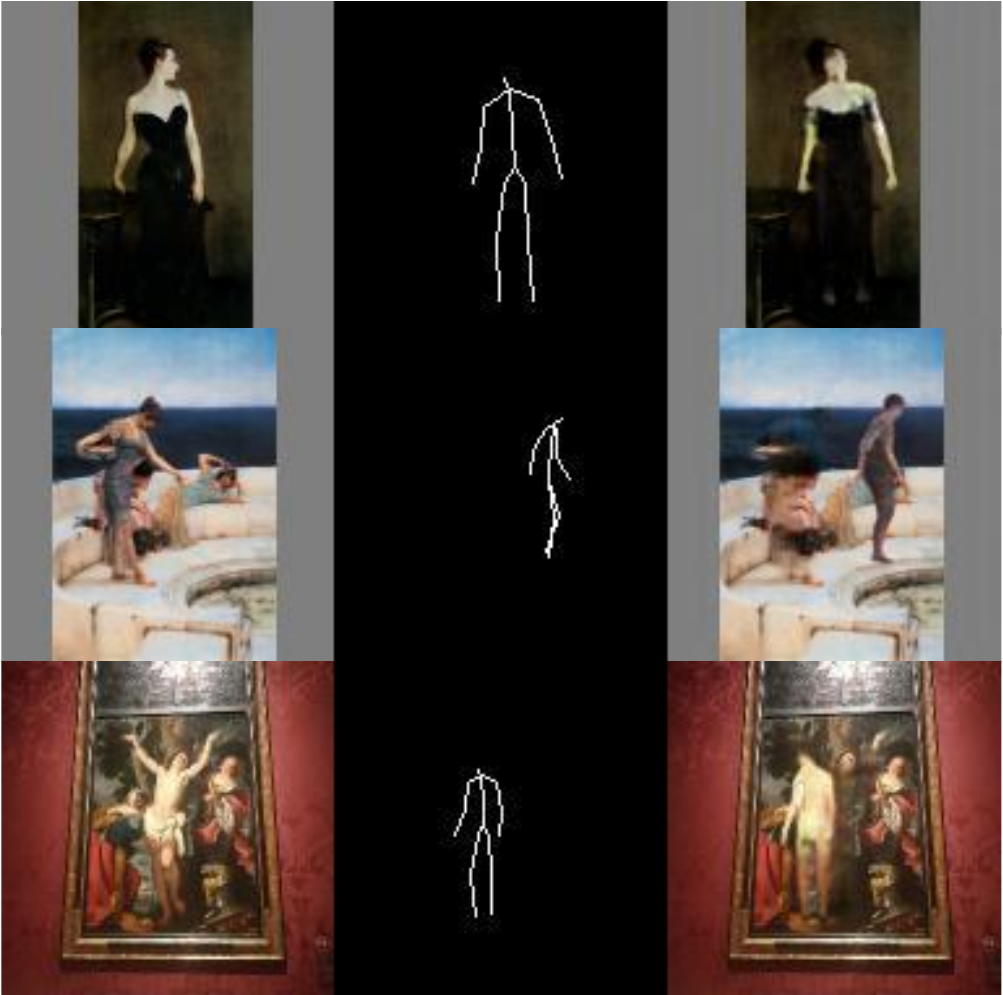}} 
\caption{ISP reposing of human figures: (a) MPII dataset \cite{andriluka20142d}, (b) LSP dataset \cite{Johnson10} and (c) art works in the following order,  Madame X (1884)--John Singer Sargent,
Silver Favourites (1903)--Lawrence Alma-Tadema,
Saint Sebastian Tended--Saint Irene and her Maid-Bernardo Strozzi.}
\label{fig:realRps_2}
\vspace{-.2in}
\end{figure}
}

\newcommand{\figWillamsRush}
{
\begin{figure}[t]
 \centering
 \subfloat[]{
 \label{fig:Willam_rush_A}
 \includegraphics[width=0.3\linewidth,{trim=0in 0in 0in 0in,clip=true}]{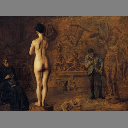}}
 \subfloat[]{
 \label{fig:Willam_rush_O1}
 \includegraphics[width=0.3\linewidth,{trim=0in 0in 0in 0in,clip=true}]{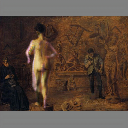}}
  \subfloat[]{
 \label{fig:Willam_rush_O2}
 \includegraphics[width=0.3\linewidth,{trim=0in 0in 0in 0in,clip=true}]{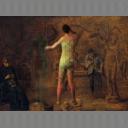}}
\caption{Inner space preserving reposing of one of Thomas Eakins' paintings: William Rush Carving His Allegorical Figure of the Schuylkill River, 1908. (a) original painting, (b) one reposed painting, (c) another repose painting. \com{can you increase the resolution of the original image? Also, could you remove the gray boarders when you display them? Also, could you display the new pose model on top right corner for each reposed image?}}
\label{fig:WillamRush}
\vspace{-.2in}
\end{figure}
}

\newcommand{\figRealRps}
{
\begin{figure}[t]
 \centering
 \subfloat[]{
 \label{fig:mpiiRps}
 \includegraphics[width=0.48\linewidth,{trim=0in 0in 0in 0in,clip=true}]{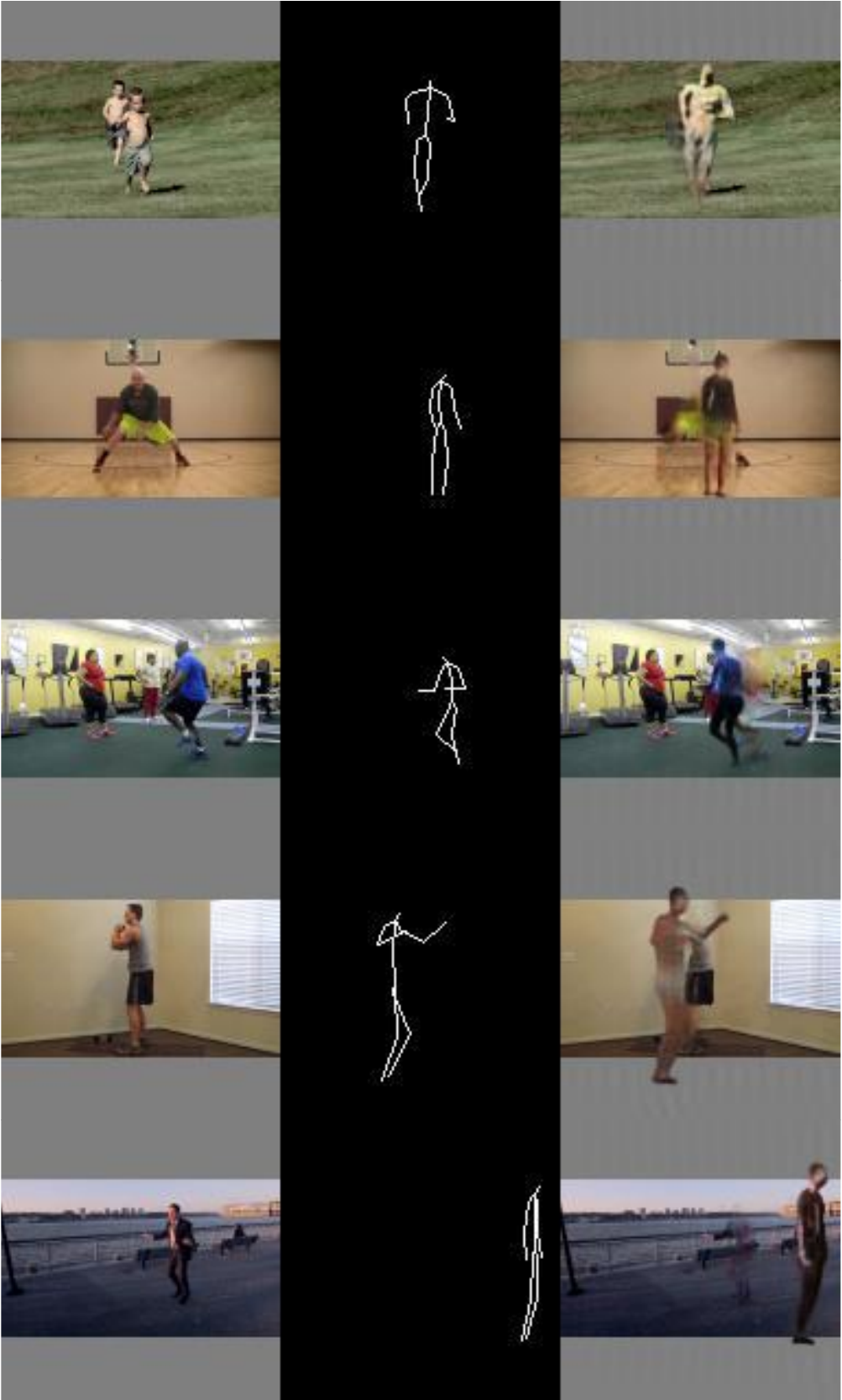}}
 \subfloat[]{\label{fig:lspRps}
 \includegraphics[width=0.48\linewidth,{trim=0in 0in 0in 0in,
  clip=true}]{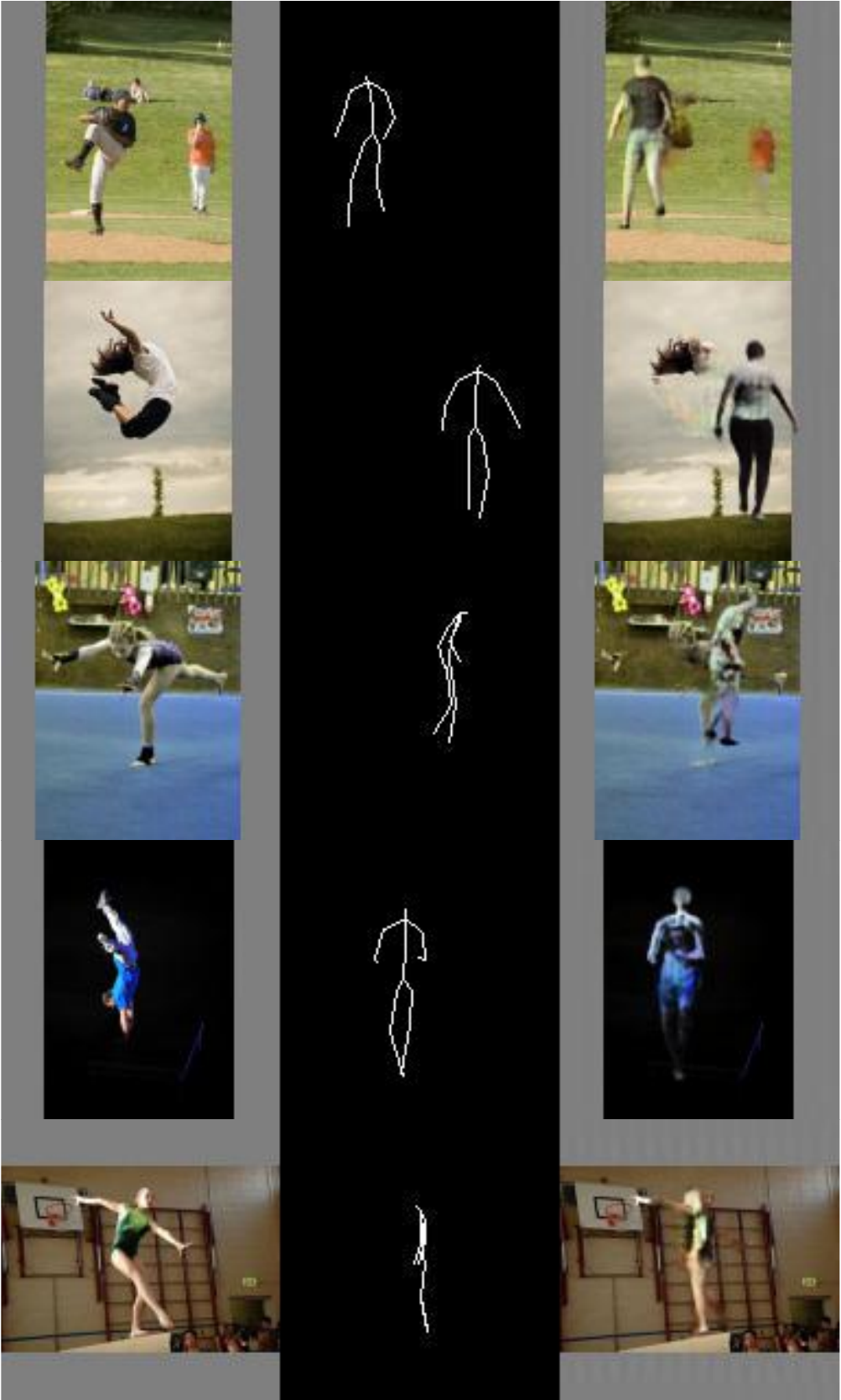}} 
\caption{ISP reposing of public human pose datasets: (a) MPII dataset \cite{andriluka20142d}, and (b) LSP dataset \cite{Johnson10}.}
\label{fig:realRps}
\vspace{-.2in}
\end{figure}
}

\newcommand{\figMaxPFCcomp}
{
\begin{figure}[t]
 \centering
 \subfloat[]{
 \label{fig:maxP_rst}
 \includegraphics[width=0.48\linewidth,{trim=0in 0in 0in 0in,clip=true}]{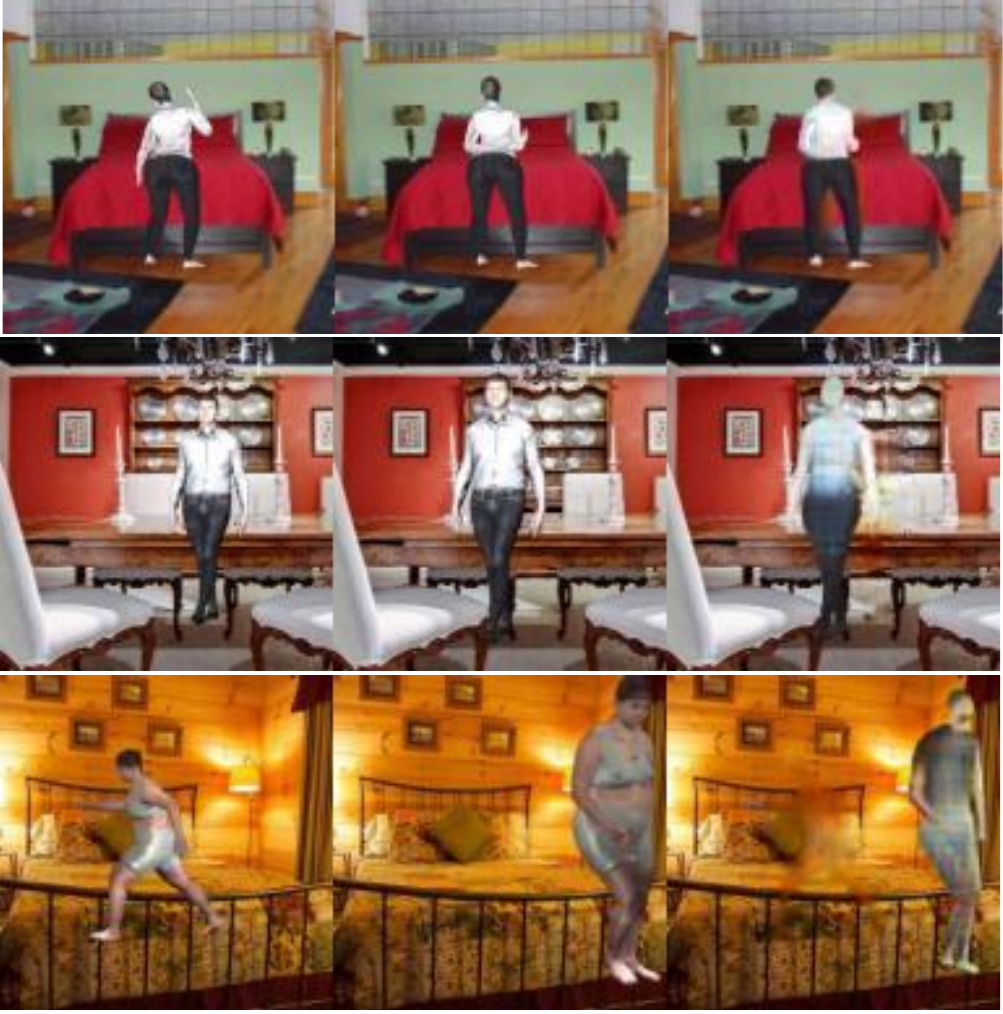}}
 \subfloat[]{
 \label{fig:FC_l2_rst}\includegraphics[width=0.48\linewidth,{trim=0in 0in 0in 0in,
  clip=true}]{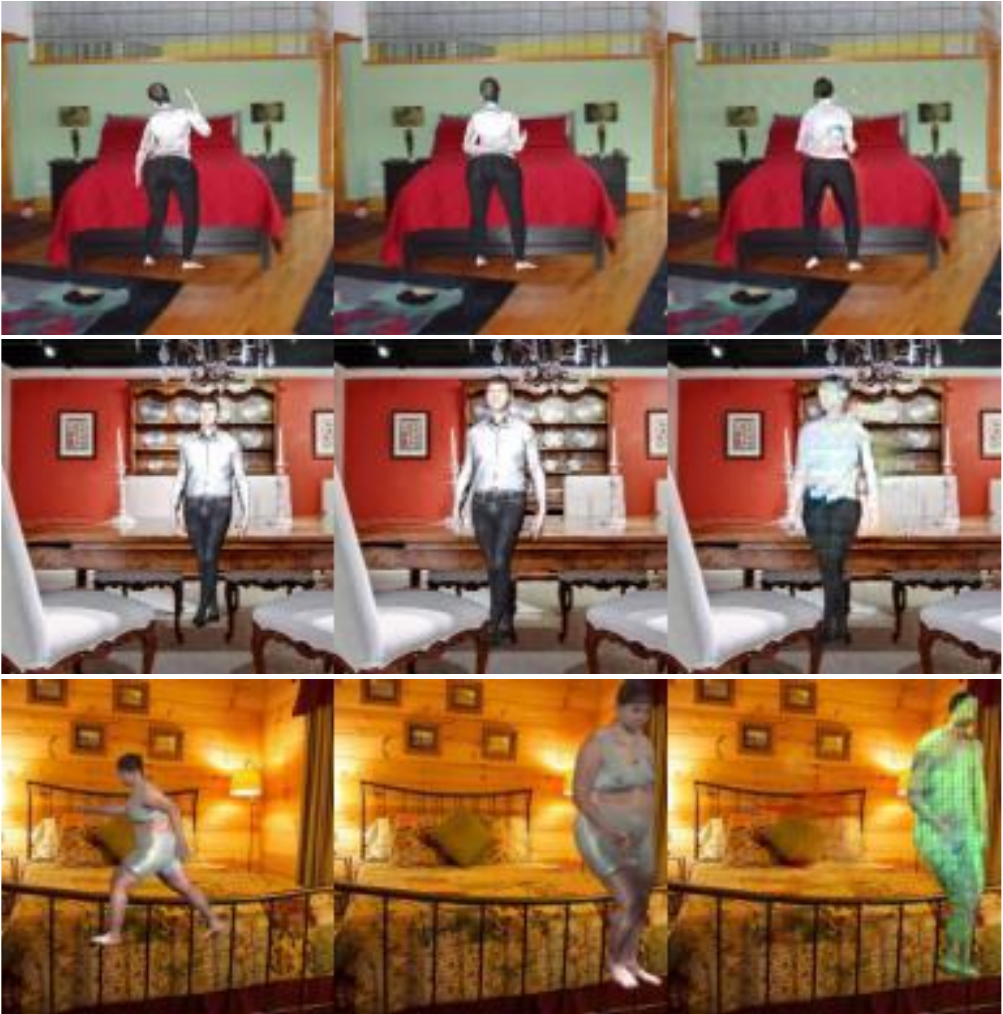}} 
\caption{Inner space preserving human reposing with different downsampling layers: (a) downsampled with max pooling, and (b) downsampled with convolution layers. First column is the input image, second column is the ground truth image of the target pose, last column is the generated image from ISP-GPM.}
\vspace{-.2in}
\label{fig:maxPFCcomp}
\end{figure}
}

\newcommand{\figDlayerLoss}
{
\begin{figure}[t]
 \centering
 \subfloat[]{
 \label{fig:Dl_l1_loss}
 \includegraphics[width=0.45\linewidth,{trim=0in 0in 0in 0in,clip=true}]{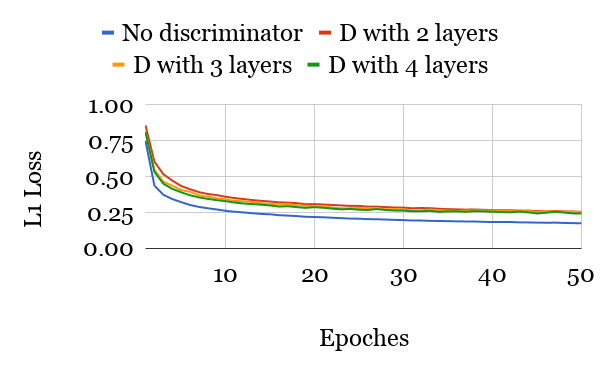}}
    \subfloat[]{
 \label{fig:Dl_G_loss}\includegraphics[width=0.45\linewidth,{trim=0in 0in 0in 0in,
  clip=true}]{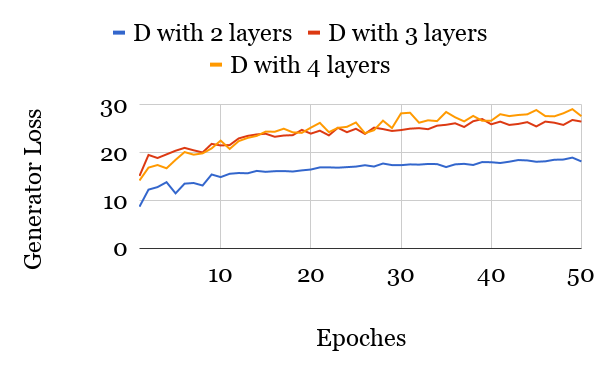}} 
\caption{Losses during training for different network configurations: (a) L1 loss, (b) Generator loss. Note that model without discriminator only shows in L1 loss.}
\label{fig:DlayerLoss}
\vspace{-.2in}
\end{figure}
}

\newcommand{\figRstComp}
{
\begin{figure}[b]
\vspace{-.2in}
 \centering
 \subfloat[]{
 \label{fig:genClothRst}
 \includegraphics[width=0.31\linewidth,{trim=0in 0in 0in 0in,clip=true}]{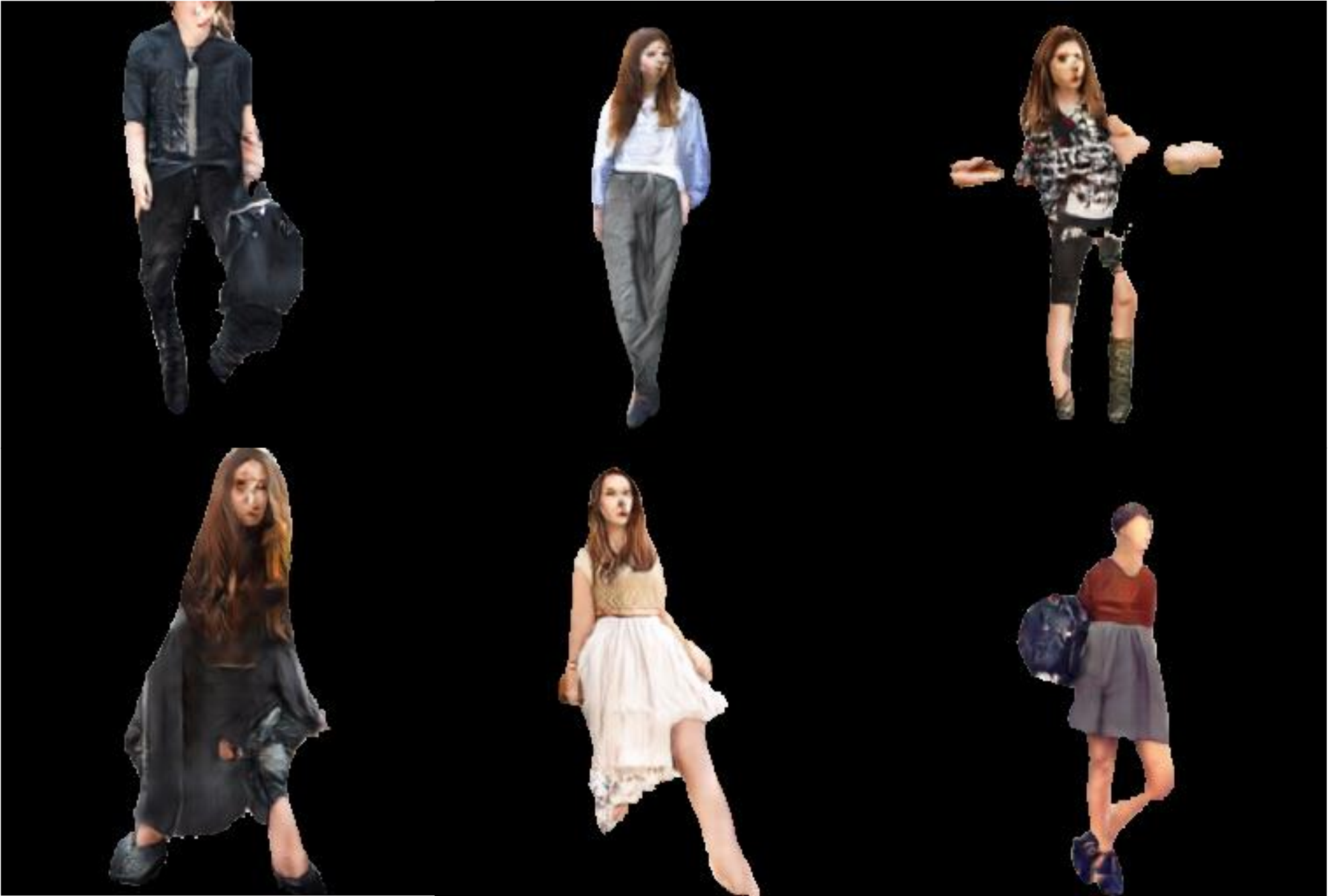}}
 \subfloat[]{
 \label{fig:LWWrst}\includegraphics[width=0.31\linewidth,{trim=0in 0in 0in 0in,
  clip=true}]{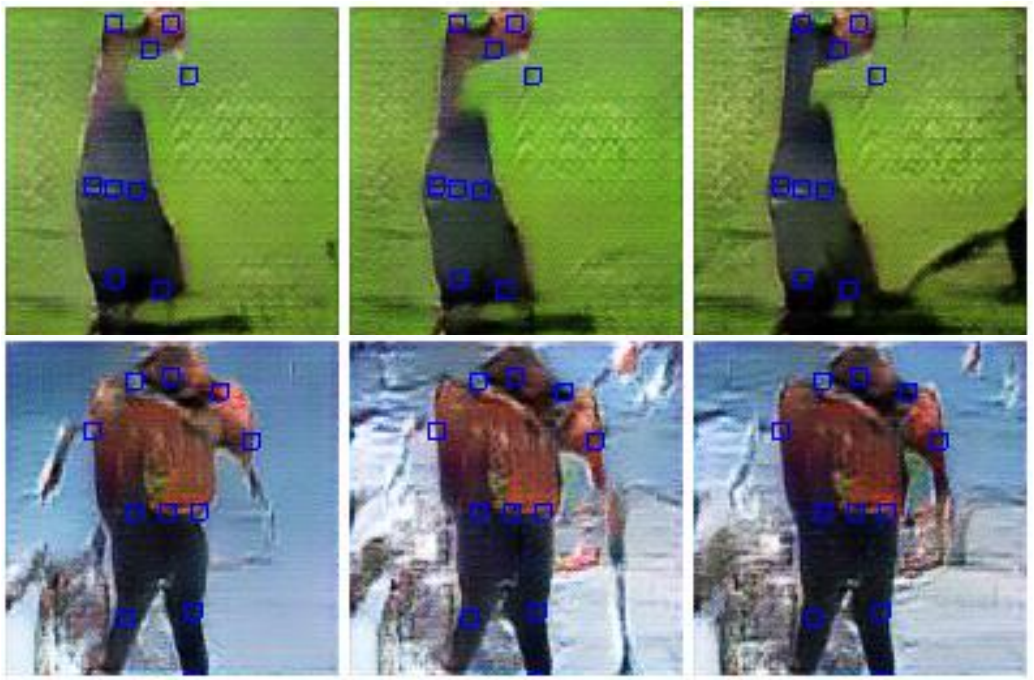}} 
   \subfloat[]{
 \label{fig:ISP-GPMrst}\includegraphics[width=0.31\linewidth,{trim=0in 0in 0in 0in,
  clip=true}]{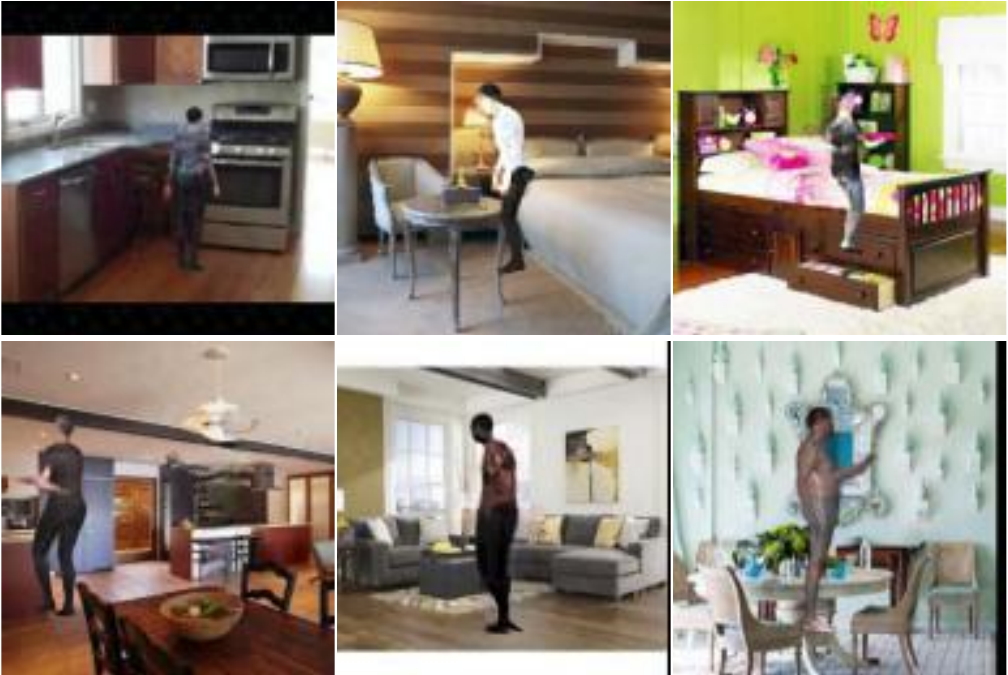}} 
\caption{Image quality comparison of the generative models for human figures presented by (a) Lassner \cite{lassner2017generative},  (b) Reed \cite{reed2016learning}, and (c) our ISP-GPM.}
\label{fig:rstComp}
\vspace{-.2in}
\end{figure}
}

\newcommand{\figresults}
{
\begin{figure}[t]
 \centering
    \subfloat[]{
 \label{fig:res_1}\includegraphics[width=0.7\linewidth,{trim=0in 0in 0in 0in,
  clip=true}]{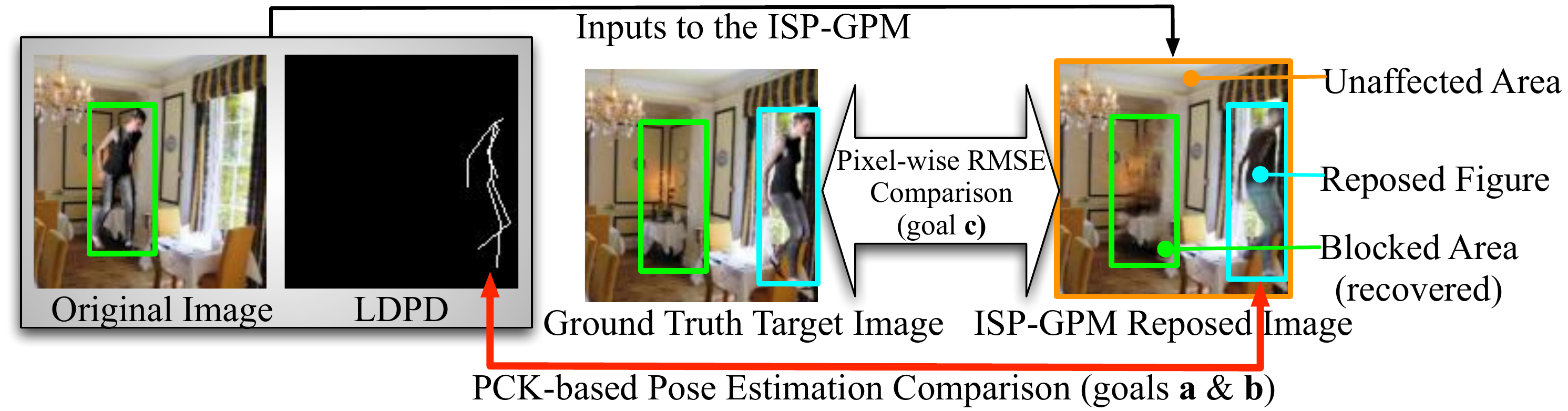}} 
  \subfloat[]{
 \label{fig:res_2}
 \includegraphics[width=0.30\linewidth,{trim=0in 0in 0in 0in,clip=true}]{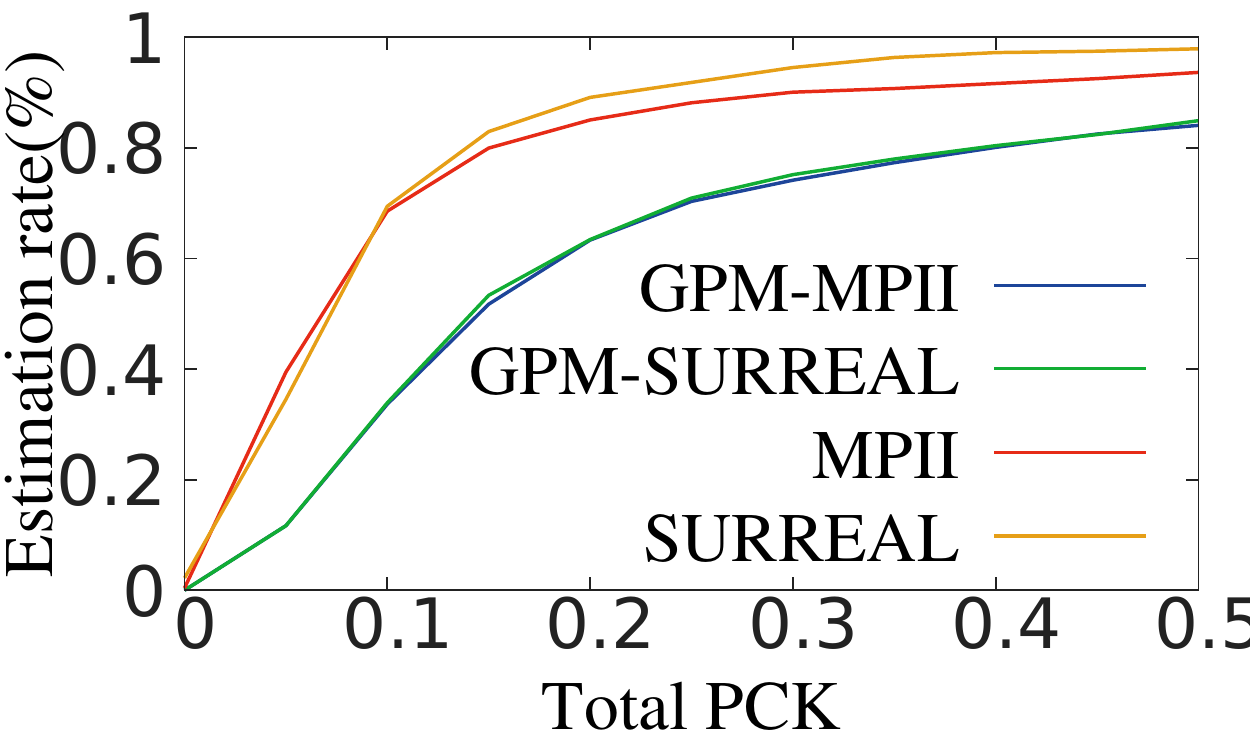}}
\caption{(a) ISP quantitative evaluation schematic, (b) Pose estimation accuracy comparison tested on MPII, SURREAL, and our ISP-GPM datasets.}
\vspace{-.2in}
\end{figure}
}
\newcommand{\tabMthComp}{
\begin{table}[h]
\scriptsize
\caption{Pose generators comparison}
\vspace{-.1in}
\begin{center}
 \begin{tabular}  { m{1.1cm}  m{1.1cm}  m{1.1cm} m{1.1cm} m{1.1cm} m{1.1cm}} 
 \hline
 Methods & Sampling-based & Instance-based & ISP & \# of Components & Articulation\\
  \hline
  Reed's \cite{reed2016generating} & $\circ$ & $\cross$ &  $\cross$ & N & $\circ$ \\
  Reed's \cite{reed2016learning} & $\circ$ & $\cross$ &  $\cross$ & N & $\circ$ \\
  Lassner's \cite{lassner2017generative}  & $\circ$ & $\cross$ &  $\cross$ & N & $\circ$ \\
 \hline
 Kulkarni's \cite{kulkarni2015deep} & $\cross$ & $\circ$ &  $\circ$ & 1 & $\cross$ \\
 Hinton \cite{hinton2011transforming} & $\cross$ & $\circ$ &  $\circ$ & 1 & $\cross$ \\
Dosovitskiy's \cite{dosovitskiy2015learning} & $\cross$ & $\circ$ &  $\circ$ & 1 & $\cross$ \\
 Dosovitskiy's \cite{dosovitskiy2017learning} & $\cross$ & $\circ$ &  $\circ$ & 1 & $\cross$ \\
\textbf{ISP-GPM} & $\cross$ & $\circ$ &  $\circ$ & N & $\circ$ \\
 \hline 
 \end{tabular} 
 \vspace{-.2in}
\label{tbl:mathComp}
\end{center}
\end{table}
}

\newcommand{\eqnref}[1]{Eq.~(\ref{eqn:#1})}
\newcommand{\figref}[1]{Fig.~\ref{fig:#1}}
\newcommand{\tblref}[1]{Table~\ref{tbl:#1}}
\newcommand{\secref}[1]{Section~\ref{sec:#1}}
\newcommand{\thmref}[1]{Theorem~\ref{thm:#1}}
\newcommand{\defref}[1]{Definition~\ref{definition:#1}}
\newcommand{\lemref}[1]{Lemma~\ref{lem:#1}}
\newcommand{\com}[1]{\textcolor{red}{#1}}
\maketitle              

\begin{abstract}
Image-based generative methods, such as generative adversarial networks (GANs) have already been able to generate realistic images with much context control, specially when they are conditioned. However, most successful frameworks share a common procedure which performs an image-to-image translation with pose of figures in the image untouched. When the objective is reposing a figure in an image while preserving the rest of the image, the state-of-the-art mainly assumes a single rigid body with simple background and limited pose shift, which can hardly be extended to the images under normal settings. In this paper, we introduce an image ``inner space'' preserving model that assigns 
an interpretable low-dimensional pose descriptor (LDPD) to an articulated figure in the image. Figure reposing is then generated by passing the LDPD and the original image through multi-stage augmented hourglass networks in a conditional GAN structure, called inner space preserving generative pose machine (ISP-GPM). We evaluated ISP-GPM on reposing human figures, which are highly articulated with versatile variations. Test of a state-of-the-art pose estimator on our reposed dataset gave an accuracy over 80\% on PCK0.5 metric.
The results also elucidated that our ISP-GPM is able to preserve the background with high accuracy while reasonably recovering the area blocked by the figure to be reposed.


\vspace{-.1in}
\end{abstract}
\vspace{-.4in}

\section{Introduction}
\vspace{-.1in}
Photographs are important because they seem to capture so much: in the right photograph we can almost feel the sunlight, smell the ocean breeze, and see the fluttering of the birds. And yet, none of this information is actually present in a two-dimensional image. Our human knowledge and prior experience allow us to recreate ``much'' of the world state (i.e. its inner space) and even fill in missing portions of occluded objects in an image since the manifold of \emph{probable} world states has a lower dimension than the world state space.

Like humans, deep networks can use context and learned ``knowledge'' to fill in missing elements. But more than that, if trained properly, they can modify (repose) a portion of the inner space while preserving the rest, allowing us to significantly change portions of the image. In this paper, we present a novel deep learning based generative model that takes an image and pose specification and creates a similar image in which a target element is reposed. In \figref{WillamsRushB}, we reposed a human figure a number of different ways based on a single painting by the early 20th century painter, Thomas Eakins.


\figWillamsRushB

In reposing a figure there are three goals: (\textbf{a}) the output image should look like a realistic image in the style of the source image, (\textbf{b}) the figure should be in the specified pose, and (\textbf{c}) the rest of the image should be as similar to the original as possible. Generative adversarial networks (GANs) \cite{goodfellow2014generative}, are the ``classic'' approach to solving the first goal by generating novel images that match a certain style. More recently, other approaches have been developed that merge deep learning and probabilistic models including the variational autoencoder (VAE) to generate realistic images \cite{rezende2014stochastic,radford2015unsupervised,larsen2016autoencoding,dosovitskiy2017learning,bao2017cvae,zhang2016colorful,lassner2017generative,pathak2016context,yi2017dualgan}. 

The second goal, putting the figure in the correct pose, requires a more controlled generation approach. Much of the work in this area is based around conditional GANs (cGAN) \cite{mirza2014conditional} or conditional VAE (cVAE) \cite{sohn2015learning,larsen2016autoencoding}. 
The contextual information can be supplied in a variety of ways. Many of these algorithms generate based on semantic meaning, which could be class labels, attributes, or text descriptors \cite{gauthier2014conditional,yan2016attribute2image,reed2016generative,walker2016uncertain,odena2016conditional}. Others are conditioned on an image often called as image-to-image translation \cite{yi2017dualgan}. The success of image-to-image translation is seen in many tasks including colorization \cite{zhang2016colorful,larsson2016learning,iizuka2016let}, semantic image segmentation \cite{chen2018deeplab,long2015fully,ronneberger2015u,hariharan2014simultaneous,mostajabi2015feedforward,dai2015convolutional,ning2005toward,farabet2013learning,pinheiro2014recurrent,ciresan2012deep}, texture transfer \cite{efros2001image}, outdoor photo generation with specific attributes  \cite{shih2013data,laffont2014transient}, 
scene generation with semantic layout \cite{karacan2016learning}, and product photo generation \cite{yoo2016pixel,eitz2012humans}. 

At a superficial level, this seems to solve the reposing problem. However, these existing approaches generally either focus on preserving the image (goal \textbf{c}) or generating an entirely novel image based on the contextual image (goal \textbf{b}), but not both. For example, when transforming a photo of a face to a sketch, the result will keep the original face spatial contour unchanged \cite{yi2017dualgan}, and when generating a map from a satellite photo, the street contours will be untouched \cite{isola2017image}. Conversely, in attribute based generation, the whole image is generated uniquely for each description \cite{yan2016attribute2image,karacan2016learning}, so even minor changes will result in completely different images. A demo case from an attribute based bird generation model from \cite{reed2016learning,reed2016generative} is demonstrated in \figref{NipsLWW}, in which only changing a bird's head color from black to red will alter nearly the entire image.\footnote{For this experiment, the random term was set to zero to rule out differences due to the input.}

Recently, there have been attempts to change some elements of the inner space while preserving the remaining elements of an image. Some works successfully preserve the object graphical identities with varying poses or lighting conditions \cite{kulkarni2014inverse,loper2014opendr,kulkarni2015deep,jampani2015informed,hinton2011transforming,michalski2014modeling,dosovitskiy2015learning,yang2015weakly}. These works include human face or office chair multi-view regeneration. Yet, all these works are conducted under simplified settings that assume a single rigid body with barren textures and no background. Another work limited the pose range to stay on the pose manifold \cite{yang2015weakly}. This makes them very limited when applied on images from natural settings with versatile textures and cluttered background.


\figNipsLWW

We address the problem of articulated figure reposing while preserving the image's inner space (goals \textbf{b} and \textbf{c}) via the introduction of our inner space preserving generative pose machine (ISP-GPM) that generates realistic reposed images (goal \textbf{a}). In ISP-GPM, an interpretable low-dimensional pose descriptor (LDPD) is assigned to the specified figure in the 2D image domain. Altering LDPD causes figure to be reposed. For image regeneration, we used stack of augmented hourglass networks in a cGAN framework, conditioned on both LDPD and the original image. We replaced hourglass network original downsampling mechanism by pure convolutional layers to maximize the  ``inner space''  preservation between the original and reposed images. Furthermore, we extended the ``pose'' concept to a more general format which is no longer a simple rotation of a single rigid body, but instead the relative relationship between all the physical entities present in an image and its background. We push the boundary to an extreme case---a highly articulated object (i.e. human body) against a naturalistic background (code available at \cite{ISPGPM}). A direct outcome of ISP-GPM is that by altering the pose state in an image, we can achieve unlimited generative reinterpretation of the original world, which ultimately leads to a one-shot ISP data augmentation. 



\section{Related Work}
Pose altering is very common in our physical world. If we take photographs of a dynamic articulated object over time, they can hardly be the same. These images share a strong similarity due to having a relatively static background with only differences caused by changes in the object's pose states. We can perceive these differences since the pose information is partially reflected in these images. However, the true ``reposing'' actually happens in the 3D space and the 2D mapping is just a simple projection afterwards. This fact inspired 3D rendering engines such as Blender, Maya, or 3DS Max to simulate the physical world in (semi)exact dimensions at graphical level, synthesize 3D objects in it, repose the object in 3D, and then finally render a 2D image from the reposed object using a virtual camera \cite{lassner2017generative}. 
Following this pipeline, there are recent attempts to generate synthesized human images \cite{pons2013metric,shotton2011real,taylor2012vitruvian}.  SCAPE method parameterizes the human body shapes into a generalized template using dense 3D scans of a person in multiple poses \cite{anguelov2005scape}. Authors in \cite{chen2018deeplab} mapped the photographs of clothing into SCAPE model to boost human 3D pose dataset.  Physical rendering and real textures are combined in \cite{varol2017learning} to generate a synthetic human dataset. However, these methods inevitably require sophisticated 3D rendering engines and avatar data is needed either from full 3D scanning with special equipment or generated from generalized templates \cite{loper2015smpl,anguelov2005scape}, which means such data is not easily accessible or extendable to novel figures. 


Image-based generative methods, such as GANs and VAEs have already been able to generate realistic images with much context control, specially when they are conditioned \cite{isola2017image,bao2017cvae,reed2016generative}. There are also works addressing pose issue of rigid (e.g. chair \cite{dosovitskiy2016generating}) or single (e.g. face \cite{yang2015weakly}) objects. An autoencoder structure to capture shift or rotation changes is employed in \cite{larsen2016autoencoding}, which successfully regenerates images of 2D digits and 3D graphics rendered images with pose shift.  Deep convolutional inverse graphics network (IGN) \cite{kulkarni2015deep} learns interpretable representation of images including out-of-plane rotations and lighting variations to generate face and chairs from different view points. Based on IGN concept, Yang employed a recurrent network to apply out-of-plane rotations to human faces and 3D chairs to generate new images \cite{yang2015weakly}.  In \cite{dosovitskiy2015learning}, authors built a convolutional neural network (CNN) model for chair view rendering, which can interpolate between given viewpoints to generate missing ones or invent new chair styles by interpolating between chairs from the training set. By incorporating 3D morphable model into a GAN structure, the authors in  \cite{yin2017towards} proposed a framework which can generate face frontalization in the wild with less training data. These works as a matter of fact in a sense preserve the inner space information with the target identity unchanged. However, most are limited to a single rigid body with simple or no background, and are inadequate to deal with complex articulated objects such as human body in a realistic background setting. 


In the last couple of years, there have been a few image-based generative models proposed for human body reposing. In \cite{reed2016learning} and \cite{reed2016generative}, by localizing exact body parts, human figures were synthesized with provided attributes. However, though pose information is provided exactly, the appearance are randomly sampled under attribute context. Lassner and colleagues in \cite{lassner2017generative} generated vivid human figures with varying poses and clothing textures by sampling from a given set of attributes. A direct result of sampling based method is a strong coupling effect between different identities in the image, in which the pose state cannot get altered without the image inner space change. 

In this paper, we focus on the same pose and reposing topics but extend them to a more general format of highly articulated object with versatile background under realistic/wild settings. We are going to preserve the original inner space of the image, while altering the pose of the an specific figure in the image. Instead of applying a large domain shift on an image such as changing the day to night, or the summer to winter, we aim to model a pose shift caused by a movement in the 3D physical world, while the inner space of the world stays identical to its version before this movement. Inspired by this idea, we present our inner space preserving generative pose machine (ISP-GPM), in which rather than attribute based sampling, we focus on specific image instances. 


\section{World State and Inner Space of An Image} \label{sec:innerSpace}
\vspace{-.1in}
``No man ever steps in the same river twice'' quoted from Heraclitus. 

Our world is dynamically changing. Taking one step forward, raising hand a little bit, moving our head to the side, all these tiny motions make us visually different from a moment ago. These changes are also dependably reflected in the photographs taken from us. In most cases, for a short period of time, we can assume such changes are purely caused by pose shift instead of characteristic changes of all related entities. Let's simply call the partial world captured by an image ``the world''. If we model the world by a set of rigid bodies, for a single rigid body without background (the assumption in the most of the state-of-the-art), the world state can be described by appearance term $\boldsymbol{\alpha}$ and the pose state $\boldsymbol{\beta}$ of the rigid body as $W_s = \{\boldsymbol{\alpha}, \boldsymbol{\beta}\}$ and the reposing process is conduced by altering $\boldsymbol{\beta}$ to a target pose $\boldsymbol{\hat\beta}$. However, real world can hardly be described by a simple rigid body, but clustered articulated rigid bodies and background. In this case, we formulate the world state as: 
\begin{equation}
    \label{eqn:worldState}
    W_s = \{\boldsymbol{\alpha_i}, \boldsymbol{\beta_i}, \phi(i,j)| i,j \in N \}.
\end{equation}
where, $N$ stands for the total number of rigid bodies in the world and $\phi(i,j)$ stands for the constraints between two rigid bodies. For example, a human has $N$ (depending on the granularity of the template that we choose) articulated limbs in which the joints between them follow the biomechanical constraints of the body. 
A pure reposing process in physical world should keep the $\boldsymbol{\alpha_i}$ terms unchanged. However, in imaging process, only part of the $\boldsymbol{\alpha_i}$ information is preserved as $\boldsymbol{\alpha_i^{in}}$ with $\boldsymbol{\alpha_i = \alpha_i^{in} + \alpha_i^{out}}$, where $\boldsymbol{\alpha_i^{out}}$ stands for the missing information in the image with respect to the physical world. 
We assume each image can partially preserved the physical world information and we call this partially preserved world state the ``inner space''. If $\boldsymbol{\alpha_i^{in}}$ and $\phi(i,j)$ term are preserved during figure $i$ reposing, we call this process ``inner space preserving''. 

Another assumption is that in the majority of cases, the foreground ($F$) and the background ($B$) should be decoupled in the image, which means if figure $i \in F$ and figure $j \in B$, the $\phi(i,j)$ is empty or vice versa. This means if a bird with black head and yellow body is the foreground, the identical bird can be in different backgrounds such as on a tree or in the sky. However, strong coupling between foreground and background is often seen in attribute-based models as shown in \figref{NipsLWW}. Instead, we designed our generative pose machine to reflect: (1) inner space preserving, and  (2) foreground and background decoupling. 

\section{ISP-GPM: Inner Space Preserving Generative Pose Machine}
\vspace{-.1in}
The ISP-GPM addresses the extensive pose transformation of articulated figures in an image through the following process: given an image with specified figure and its interpretable low-dimensional pose descriptor (LDPD), ISP-GPM  outputs a reposed figure with original image inner space preserved (see \figref{GPMframe}). The key components of the ISP-GPM are: (1) a CNN interface converter to make the LDPD compatible with the first convolutional layer of the ISP-GPM interface, and (2) a generative pose machine to generate reposed figures using the regression structure of hourglass networks when stacked in a cGAN framework in order to force the pose descriptor into the regenerated images.

\figGPMframe

\subsection{CNN Interface Converter} \label{sec:CNNinter}
\vspace{-.1in}
We employed an LDPD in the 2D image domain, which in the majority of the human pose dataset such as Max Planck institute informatics (MPII) \cite{andriluka20142d} and Leeds sports pose (LSP) \cite{Johnson10} is defined as the vector of  2D joint position coordinates. To make this descriptor compatible with the convolutional layer interface of ISP-GPM, we need a CNN interface converter. The most straight forward converter could simply set the joint point in the image, similar to the work described in \cite{reed2016learning}. As human body can be represented by a connected graph \cite{andriluka2009pictorial,bergtholdt2010study}, more specifically a tree structure, in this work we further appended the edge information into our converter. 
Assume human pose to be represented by 2D locations of its $N$ joints. Let's use $N$ channel maps to hold this information as joint map, $J_{Map}$. For each joint $i$ with coordinates $(x_i, y_i)$, if joint $i$'s parent joint exists, we are going to draw a line from $(x_i,y_i)$ to its parent location in channel $i$ of $J_{Map}$. 
In generating $J_{Map}$s, the draw operation is conducted by image libraries such as OpenCV \cite{bradski2000opencv}. 


\subsection{Stacked Fully Convolutional Hourglass cGAN}
\vspace{-.1in}
Many previous works have proved the effectiveness of multi-stage estimation structure in human pose estimation, such as 2016 revolutionary work of convolutional pose machine \cite{wei2016convolutional}. As an inverse operation to regenerate figures of humans, we employed a similar multi-stage structure. Furthermore, human pose can be described in a  multi-scale fashion, starting from simple joint description to  sophisticated clothing textures on each body part, which inspired the use of an hourglass model with a stacked regression structure \cite{newell2016stacked}. However, instead of pose estimation or segmentation, for human reposing problem, more detailed information needs to be preserved in both encoding and decoding phases of the hourglass network. Therefore, we replaced hourglass network's max pooling and the nearest upsampling modules by pure convolutional layers to maximize the information preservation. The skip structure of the original hourglass network is also preserved to let more original high frequency parts pass through. Original hourglass is designed for image regression purpose. In our case, we augment hourglass original design by introducing structure losses 
 \cite{isola2017image}, which penalize the joint configuration of the output. We forced the pose into the generated image by employing a cGAN mechanism.





\figFChourglassCGAN

An overview of our stacked fully convolutional hourglass cGAN (FC-hourglass-cGAN) is shown in \figref{FChourglassCGAN}, where we employed a dual skip mechanism, a module level skip as well as the inner module level skips. Each FC-hourglass employs a encoder-decoder like structure \cite{noh2015learning,badrinarayanan2017segnet,newell2016stacked}. Stacked FC-hourglass plays the generator role in our design, while another convolutional net plays the discriminator role. We employed an intermediate supervision mechanism similar to \cite{newell2016stacked}, however the supervision is conducted by both L1 loss and generator loss, as described in the following section.

\subsection{Stacked Generator and Discriminator Losses}

Due to the ISP-GPM stacked structure, the generator loss comes from all intermediate stages to the final one. The loss for generator is then computed as:
\vspace{-.1in}
\begin{equation} \label{eqn:cGAN_loss}
    L_{G}(G,D)= \mathbb{E}_{u,v}[\log D(u,v)] +\sum_{i=1}^{N_{stk}} \mathbb{E}_{u}[\log(1-D(u,G(u)[i])].
\end{equation}
where, $u$ stands for the combined input of $J_{Map}$  and the original image, and $v$ is the target reposed image. $G$ is stacked FC-hourglass that acts as the generator role, $N_{stk}$ stands for the total number of stacks in the generator $G$,  and $D$ is the discriminator part of the cGAN. Different from commonly used generator, our $G$ gives multiple output according to the stack number. $G(u)[i]$ stands for the $i$-th output conditioned on $u$. Another difference from traditional cGAN design is that we do not include the random term $z$ as it is common in most GAN based models \cite{mirza2014conditional,sohn2015learning,gauthier2014conditional,yan2016attribute2image,odena2016conditional,goodfellow2014generative}. The particular reason to have this term in traditional GAN based model is to introduce higher variation into the sampling process. 
The main reason behind introducing randomness in GAN is to capture a probabilistic distribution which generates \emph{novel} images that match a certain style. However, our ISP-GPM follows quite opposite approach, and aims to achieve a deterministic solution based on the inner space parameters, instead of generating images from a sampling process.
$D$ term is the discriminator to reveal if the input is real or fake, conditioned on our input $u$ information. 

Since our aim is regressing the figure to a target pose on its subspace manifold, low frequency components play an import role here to roughly localize the figure to the correct position. Therefore, we capture these components using a classical L1 loss:

\vspace{-.3in}
\begin{align}  \label{eqn:losses}
    L_{L1}(G) = \sum_{i=1}^{N_{stk}} \mathbb{E}_{u,v}[||v- G(u)[i]||_1].
\end{align}

We used a weighted term $\lambda$ to balance the importance of L1 and  $G$ losses in our target objective function: 

\vspace{-.1in}
\begin{equation}
\label{eqn:lossAll}
    L^*_{obj} = \arg\,\min_G \, \max_D\, L_{G}(G,D)+ \lambda L_{L1}(G). 
\end{equation}





\section{Model Evaluation}
\vspace{-.1in}
To illustrate our inner space preserving concept and the performance of the proposed ISP-GPM, we chose a specific figure as our reposing target, the human body, due to the following rationale. First and foremost, human body is a highly articulated object with over 14 components depending on the defined limb granularity. Secondly, human pose estimation and tracking is a well-studied topic \cite{sapp2013modec,felzenszwalb2008discriminatively,wei2016convolutional,pishchulin2013strong,bourdev2009poselets,ramanan2007learning} as it is highly needed in abundant applications such as pedestrian detection, surveillance, self-driving cars, human-machine interaction, healthcare, etc. Lastly, several open-source  datasets are available including MPII \cite{andriluka20142d}, BUFFY \cite{ferrari2008progressive}, LSP \cite{Johnson10}, FLIC \cite{sapp2013modec}, and SURREAL \cite{varol2017learning}, which can facilitate deep learning-based model training and wide range of test samples for model evaluation. 

\subsection{Dataset Description}  
\vspace{-.1in}
Although well-known datasets for human pose estimation \cite{andriluka20142d,Johnson10,sapp2013modec} exist, few of them can satisfy our reposing purpose. As mentioned in \secref{innerSpace}, we aim at preserving the inner space of the original image before figure reposing. Therefor, we need pairs of images with the same $\boldsymbol{\alpha}$ term but varying $\boldsymbol{\beta}$ term, which means identical background and human. The majority of the existing datasets are collected from different people individually with no connections between images, so they have varying $\boldsymbol{\alpha}$ and $\boldsymbol{\beta}$. A better option is extracting images from consecutive frames of a video. However, not many labelled video datasets from human are available. Motion capture system can facilitate auto labeling process, but they focus on the pose data without specifically augmenting the appearance $\boldsymbol{\alpha}$, such that ``the same person may appear under more than one subject number'' as they mentioned in \cite{mocap}. The motion capture marks are also uncommon in images taken from natural settings. Another issue with daily video clips is that the background is unconstrained as it could be dynamic caused by camera motion or other independent entities in the background. Although, our framework can handle such cases by expanding world state in \eqnref{worldState} to accommodate several dynamic figures in the scene, in this paper, we focus on a case with images from a human as the figure of interest in a static yet busy background.  

Alternatively, we shift our attention to the synthesized datasets of human poses with perfect joint labeling and background control. We employed SURREAL (Synthetic hUmans foR REAL tasks) dataset of synthesized humans with various appearance textures and background \cite{varol2017learning}. All pose data are originated from the Carnegie Mellon University motion capture (mocap) dataset \cite{mocap}. The total number of video clips for training is 54265 with combined different overlap settings \cite{varol2017learning}. Another group of 504 clips are used for model evaluation. One major issue of using SURREAL to suit our purpose is that the human subjects are not always shown in the video since it employs a fixed camera setting and the subjects are faithfully driven by the motion capture data.  We filtered the SURREAL dataset to get rid of the frames without the human in them and also the clips with too short duration such as 1 frame clips.


\subsection{ISP-GPM Implementation}
\vspace{-.1in}
Our pipeline was implemented in Torch with environment settings of CUDA8.0, CUDNN 5 with NVIDIA GeForce GTX 1080-Ti. Our implementation builds on the architecture of the original hourglass \cite{newell2016stacked,varol2017learning}. Discriminator net follows the design in \cite{isola2017image}. Adams optimizer with $\beta1 = 0.5$ and learning rate of 0.0002 was employed during training \cite{DBLP:journals/corr/KingmaB14}.
We used 3 stacked hourglass with input resolution of 128$\times$128. In each hourglass, 5 convolutions configuration is employed with lowest resolution of 4$\times$4. There are skip layers at all scale levels. 

We used the weighted sum loss during generator training with more emphasis on L1 loss to give priority to the major structure generation instead of textures. We set $\lambda = 100$ in \eqnref{lossAll} as we observed transparency in the resultant image if we give a small $\lambda$. Our input is set to $128\times128 \times3$ due to the memory limitations. The pose data is $16\times2$ vector to indicate 16 key point positions of human body as defined in SURREAL dataset \cite{varol2017learning}. In training session, we employed a batch size of 3, epoch number of 5000, and conduct 50 epochs for each test.

\figMaxPFCcomp
\subsection{ISP-GPM with Different Configurations}
\label{sec:diffconf}
\vspace{-.1in}
To compare the quality of the resultant reposed images between ISP-GPMs with different model configurations, we fixed the input image to be the first frame of each test clip and the 60th or the last frame as the target pose image. 

\vspace{-.2in}
\subsubsection{Downsampling Strategies:} We first compared the quality of the reposing when fully convolution (FC) layers vs. max pooling downsampling is used in the stacked hourglass network. To make a clear comparison, we chose same test case for different model configurations and presented the input images, ground truth and generated images in \figref{maxPFCcomp}. Each row shows a test example. Columns from left to right stand for the input image, ground truth and generated result. With the given two examples, it is clear that the max pooling is prone to the blurriness, while the FC configuration outputs more detailed textures. However, the last row of \figref{maxPFCcomp} uncovers that FC configuration is more likely to result in abnormal colors when compared to the max pooling configuration. This is expectable since the max pooling prefers to preserve the local information of an area. 

\figDlayersLabs
\vspace{-.2in}
\subsubsection{Discriminator Layer:} Inspired by \cite{isola2017image}, we employed the discriminator layer with different patch sizes to test its performance. Patch sizes can be tuned by altering the discriminator layer numbers to cover patches with different sizes. 
In this experiment, all the configurations we chose can effectively generate human contours at indicated position but only differs in the image quality. So we only show the outcomes by changing the discriminator layer from two to four as depicted in 1st to 3rd row of \figref{DlayersLabs}, respectively. The figure's last row shows the output without discriminator layer. 
We discover that the discriminator did help in texture generation, however larger patches in contrast will result in strong artifacts 
as shown in 2nd and 3rd row of \figref{DlayersLabs}. In the case with no discriminator and only L1 loss, the output is obviously prone to blurriness which is consistent with findings from previous works \cite{larsen2016autoencoding,pathak2016context,isola2017image}. We believe larger patch takes higher level structure information into consideration, however the local textures on the generated human can provides better visual quality, as seen in the 1st row of \figref{DlayersLabs}) with two layers discriminator. 

\figDlayerLoss

To better illustrate the discriminator's role during training session, we recorded loss of each component during training with different network configurations as shown in \figref{DlayerLoss}. Model without discriminator are only shown in \figref{Dl_l1_loss}. Though model without discriminator shows better performance on L1 metric, it does not always yield good looking images as it prefers to pick median values among possible colors to achieve better L1. 
There are a common trend that all $G$ loss increase as training went on and the final $G$ loss is even stronger than initial state. By observing the training process, we found out it is a process that the original human start fading away while the target posed human reveals itself gradually. Indeed, no matter how strong the generator is, its output cannot be as real as original one. So, at the beginning the generated image will be more likely to fool the discriminator as it keeps much of the real image information with less artifact.

\figRstComp 
\subsection{Comparison with the State-of-the-art}
\vspace{-.1in}
There are few works focusing on human image generation via generative models, including Reed's \cite{reed2016generating,reed2016learning} and Lassner's \cite{lassner2017generative}. We compared the outputs of our ISP-GPM model with these works as shown in \figref{rstComp} (excluding \cite{reed2016generating} since the code is not provided). We omitted the input images in \figref{rstComp} and only displayed the reposed ones to provide a direct visual comparison with other methods.

\figref{rstComp} shows that Lassner's \cite{lassner2017generative} method preserves the best texture information in the generated images. However, there are three aspects in Lassner's that need to be noted. First of all, their generation process is more like a random sampling process from the human image manifold. Secondly, to condition this model on pose, SMPL model is needed for silhouette generation, which inevitably takes advantages of a 3D engine. Thirdly, they can generate humans with vivid background, however it is like a direct mask overlapping process with fully observed background images in advance \cite{lassner2017generative}. In our ISP-GPM, both human and background are generated and merged in the same pipeline. Our pose information is a low-dimensional pose descriptor that can be generated manually. Additionally, both human and background are only partially observed due to human facing direction and the occlusion caused by the human in the scene. As for \cite{reed2016learning}, the work is not an ISP model, as illustrated by an example earlier in \figref{NipsLWW}.


\figresults
\subsection{Quantitative Evaluation}
\vspace{-.1in}
To jointly evaluate goals \textbf{a} and \textbf{b}, we \emph{hypothesized} that if the generated reposed images are realistic enough with specified pose, their pose should be recognizable by a pose recognition model trained on real-world images. We employed a high performance pose estimation model with a convolutional network architecture \cite{newell2016stacked}, 
to compare the estimated pose in the reposed synthetic image against the LDPD assigned to it in the input. We selected 100 images from both \emph{MPII Human Pose} and \emph{SURREAL} datasets in continuous order to avoid possible cherry picking. We selected the 20th frame of random video sequences to repose original images to form re-rendered ISP-GPM version datasets, namely MPII-GPM and SURREAL-GPM with joint labels compatible with the MPII joint definition. Please note that to synthesize the reposed images, we used ISP-GPM model with three layers discriminator and L1 loss as described in \secref{diffconf}.

We used probability of correct keypoint (PCK) criteria for pose estimation performance evaluation, which is the measure of joint localization accuracy \cite{yang2013articulated}. The average pose estimation rates (over 12 body joints) tested on MPII-GPM and SURREAL-GPM datasets are shown in \figref{res_2} and compared with the the pose estimator accuracy \cite{newell2016stacked} tested on 100 images from original MPII and SURREAL datasets. These results illustrate that a well-trained pose estimator model is able to recognize the pose of our reposed images with over 80\% accuracy on PCK0.5 metric. Therefore, ISP-GPM not only reposes the human figure accurately, but also makes it realistic enough to fool a state-of-the-art pose detection model to take its parts as human limbs.





With respect to goal \textbf{c}, we tested the inner space preserving ability in two folds: (1) the background of the reposed image (i.e. the unaffected area) should stay as similar as possible to the original image, and (2) the blocked area by the figure in original pose should be recovered with respect to the context. To test (1), we blocked out the affected areas where the figure of interest occupies in original and target images and computed the pixel-wise mean RMSE between the unaffected area of both images
(\textbf{RMSE = 0.050$\pm$0.001}). To evaluate (2), we compared the recovered blocked area with the ground truth target image (\textbf{RMSE = 0.172$\pm$0.010}). These results elucidate that our ISP-GPM is able to preserve the background with high accuracy while recovering the blocked area reasonably. Please note that the model has never seen behind the human in the original images and it attempts to reconstruct a texture compatible with the rest of the image, hence the higher RMSE.

\figRealRpsB
\section{ISP-GPM in Real World}
\vspace{-.1in}
To better illustrate the capability of ISP-GPM, we applied it on real world images from well-known datasets, MPII \cite{andriluka20142d} and LSP \cite{Johnson10}. As there is no ground truth to illustrate the target pose, we visualized the LDPD into a skeleton image by connecting the joints according to their kinematic relationships. ISP reposed images of MPII \cite{andriluka20142d} and LSP \cite{Johnson10} are shown in \figref{mpiiRps_2} and \figref{lspRps_2}, respectively. Each sample shows input image, visualized skeleton, and the generated image from left to right.

Arts are originated from real world and we believe when created, they also preserved inner space of an imagined world by the artist. So, we also applied our ISP-GPM on the arts inspired by human figures including paintings and sculptures. They are either from publicly accessible websites or art works in museums captured by a regular smartphone camera. The ISP reposing results are shown in \figref{artsRps_2}. From results of the real world images, the promising performance of ISP-GPM is apparent. However, there are still failure cases such as the residue of the original human that the network is unable to fully erased or the loss of the detailed texture and shape information. 



\bibliographystyle{splncs04}
\bibliography{ref}
\end{document}